\documentclass{article}

\usepackage{PRIMEarxiv}

\usepackage[utf8]{inputenc} 
\usepackage[T1]{fontenc}    
\usepackage{hyperref}       
\usepackage{url}            
\usepackage{booktabs}       
\usepackage{amsfonts}       
\usepackage{nicefrac}       
\usepackage{microtype}      
\usepackage{lipsum}
\usepackage{fancyhdr}       
\usepackage{graphicx}
\usepackage{adjustbox}
\graphicspath{{media/}}     
\usepackage{setspace}
\title{GESH-Net: Graph-Enhanced Spherical Harmonic Convolutional Networks for Cortical Surface Registration}

\author{
    Ruoyu Zhang\textsuperscript{1,†}, Lihui Wang\textsuperscript{1,*}, Kun Tang\textsuperscript{1}, Jingwen Xu\textsuperscript{1}, Hongjiang Wei\textsuperscript{2}
}

\date{}

\begin{document}
\onehalfspacing  
\maketitle


\footnotetext[1]{Engineering Research Center of Text Computing \& Cognitive Intelligence, Ministry of Education, Key Laboratory of Intelligent Medical Image Analysis and Precise Diagnosis of Guizhou Province, State Key Laboratory of Public Big Data, College of Computer Science and Technology, Guizhou University, Guiyang, 550025, China}
\footnotetext[2]{School of Biomedical Engineering, Shanghai Jiao Tong University, Shanghai, China}
\setcounter{footnote}{0}  
\renewcommand{\thefootnote}{\fnsymbol{footnote}}  
\footnotetext[1]{Corresponding author: \texttt{lhwang2@gzu.edu.cn}}
\footnotetext[2]{First author: \texttt{ZRY971123@gmail.com}}
\renewcommand{\thefootnote}{\arabic{footnote}}  

\begin{abstract}
Cortical surface registration plays a crucial role in coordinating individual cortical functions and anatomical features, serving as a fundamental step in cortical surface analysis. Its aim is to align the anatomical or functional regions of different individuals, which is of great importance for neuroimaging studies across different populations. Currently, cortical surface registration techniques based on classical methods have been well developed. However, a key issue with classical registration methods is that for each pair of images to be registered, it is necessary to search for the optimal transformation in the deformation space according to a specific optimization algorithm until the similarity measure function converges, which cannot meet the requirements of real-time and high-precision in medical image registration. With the spectacular success of deep learning in the field of computer vision, researching cortical surface image registration techniques based on deep learning models has become a new direction. But so far, there are still only a few studies on cortical surface image registration based on deep learning. Moreover, although deep learning methods theoretically have stronger representation capabilities, surpassing the most advanced classical methods in registration accuracy and distortion control remains a challenge. Therefore, to address this challenge, this paper constructs a deep learning model to study the technology of cortical surface image registration. The specific work is as follows:
(1) An unsupervised cortical surface registration network based on a multi-scale cascaded structure is designed, and a convolution method based on spherical harmonic transformation is introduced to register cortical surface data. This solves the problem of scale-inflexibility of spherical feature transformation and optimizes the multi-scale registration process. The results show that the proposed network outperforms the other deep learning-based registration methods and most classical methods, especially in terms of registration efficiency.
(2) By integrating the attention mechanism, a graph-enhenced module is introduced into the registration network, using the graph attention module to help the network learn global features of cortical surface data, enhancing the learning ability of the network. The results show that the graph attention module effectively enhances the network's ability to extract global features, and its registration results have significant advantages over other methods. Although the time consumption is slightly increased, it is still superior to other deep learning methods and much faster than classical methods

\end{abstract}

\keywords{Medical image registration \and Cortical surface registration \and Spherical convolutional neural network \and Spherical harmonic transformation \and Cascaded network \and Graph attention}

\section{Introduction}
Cortical surface registration, as a critical step in surface-based analysis, plays an essential role in neuroimaging processing. Cortical surface registration aims to align brain images from different individuals, time points, or modalities into a standard space, enabling quantitative analysis and cross-individual comparisons\cite{Ou2014}. This technique is widely applied in clinical and research fields, including neuroanatomy, neurology, brain functionality, and neuroimaging. Through cortical surface registration, researchers can better understand individual differences in brain structure and function, shedding light on changes during neurodevelopment, aging, and disease progression\cite{Mills2014}.The complexity of cortical data requires a significant amount of processing time for its registration\cite{Toga2001}, especially for nonlinear registration, like FreeSurfer\cite{Fischl2012} requiring more than 30 minutes for processing. This inefficiency, especially when dealing with large-scale datasets, poses challenges for downstream analyses such as anatomical parcellation and group-level analysis.

Classical cortical registration methods are based on the spherical topology naturally exhibited by the cortex, adopting a strategy that minimizes the deformation of triangle areas and angles to map the complex cortical data onto a sphere, where registration is performed. These methods formulate the registration process as an optimization problem in spherical space, aligning similar features by moving vertices and implementing smoothness constraints on the sphere's deformation to maintain topological integrity. Representative methods based on this idea include FreeSurfer\cite{Fischl2012}, Spherical Demons (SD)\cite{Yeo2010}, and Multimodal Surface Matching (MSM)\cite{Robinson2014,Robinson2018}.FreeSurfer\cite{Fischl2012} is a widely used neuroimaging analysis tool, whose surface registration method particularly focuses on reducing the geometric differences between the source and target cortices. By adopting a gradient descent optimization strategy, FreeSurfer minimizes the mean square error (MSE) between the source and target cortical feature maps, effectively estimating the vertex deformation field. This approach is suitable for registering geometric features such as average convexity and mean curvature, with good stability and a wide range of applications. However, it also faces challenges when handling multi-dimensional feature data. Additionally, FreeSurfer's computation speed is relatively slow, taking over an hour to complete the registration of a set of cortical surfaces, which limits its use for large-scale datasets.To improve registration efficiency, SD\cite{Yeo2010} extended the basic principles of FreeSurfer by introducing the Demons algorithm\cite{Vercauteren2009} and Gauss-Newton methods\cite{Wang2012}, significantly enhancing computational efficiency. SD emphasizes the application of the diffeomorphism concept in spherical space, ensuring the integrity of the cortical surface topology during registration and effectively avoiding structural self-intersection issues. This approach is particularly effective for applications requiring high efficiency while maintaining topological invariance. However, SD still cannot overcome the disadvantage of multi-dimensional feature registration.Recent studies\cite{Robinson2014,Glasser2016,Coalson2018} have shown that using only geometric features for registration does not yield optimal correspondence in cortical functional areas, as folding features are not the only criteria for processing cortical data. As a result, some studies\cite{Gopinath2019,Wu2019} have attempted to use spectral embedding features, functional activation features, or connectivity features for cortical surface registration. However, due to their specific mathematical modeling, these methods are still difficult to extend to multi-dimensional or high-dimensional features.To extend cortical registration to multi-dimensional features, MSM\cite{Robinson2014,Robinson2018} first proposed modeling the registration problem as a discrete labeling problem to handle multimodal features, thus providing flexibility in selecting feature sets and similarity measures. However, discrete registration method introduces higher computational overhead and presents limitations in achieving diffeomorphism. Although MSM avoids topological errors to some extent by limiting the size of deformations, it also struggles to provide optimal deformation solutions when handling large deformations.Despite the widespread deployment of the above classical methods in the neuroimaging field, they share a common major problem—high computational burden and long processing times, limiting their application in large-scale neuroimaging studies. With increasing demands for precision and efficiency, developing new algorithms and optimizing existing methods to reduce computational complexity and improve registration speed has become a significant research direction in the field of cortical registration.

In recent years, deep learning-based methods have become a frontier in cortical surface registration research due to their outstanding computational efficiency and registration performance. These methods\cite{Zhao2021,Suliman2022,Cheng2020,Suliman2023} have significantly improved the computational efficiency of cortical registration while achieving comparable registration performance to classical approaches.To accommodate a series of operations in deep learning, such as convolution, pooling, and sampling, deep learning methods need to project cortical data onto a more regular structure based on the spherical projection used in classical methods. Once the registration is completed, the deformation is mapped back to the original spherical surface. Representative works based on this approach include SphereMorph\cite{Cheng2020}, S3Reg\cite{Zhao2021}, Deep-Discrete Learning Framework (DDR)\cite{Suliman2022}, GeoMorph\cite{Suliman2023}, and SUGAR\cite{REN2024103122}. SphereMorph parameterizes the spherical mesh as a 2D rectangular image through planar projection and uses classical convolutional neural networks\cite{Keiron} to process the projected rectangular image, achieving registration accuracy comparable to FreeSurfer and SD, with a processing time of less than one minute. However, the 2D projection and inverse projection used in SphereMorph introduce additional topological errors and distortions.To avoid the issues caused by 2D projections, S3Reg, DDR, and GeoMorph project cortical data onto a regular icosahedral sphere\cite{Praun2003} for processing, thereby preserving the topology of the cortex. These methods perform convolution operations directly on the sphere, constructing spherical convolutional neural networks. S3Reg uses hexagonal filters to process spherical features, achieving accuracy comparable to some classical methods while further improving computational speed, with processing time under 10 seconds. However, due to the lack of a global coordinate system, the local coordinates of the hexagonal filters used in S3Reg flip at the spherical poles, leading to pole distortions. Although S3Reg designed a registration method that fuses three orthogonal spheres to avoid pole distortions as much as possible, the fusion also introduces new computational losses.
To address the pole distortion problem, DDR integrates MoNet\cite{Fawaz2021} into the spherical convolutional neural network. MoNet is a general graph neural network (GNN) model\cite{Zhou2020} suitable for non-Euclidean structured data, which is rotation-equivariant on the sphere, thus avoiding the pole distortion problem\cite{Suliman2022}. This gives it better registration performance compared to S3Reg, and since it does not require 2D projection or tri-orthogonal sphere fusion, DDR has higher computational efficiency than SphereMorph. Unlike the continuous registration frameworks of SphereMorph and S3Reg, DDR adopts a discrete registration framework\cite{Glocker2011} based on the idea of MSM, which makes DDR better suited for multi-dimensional feature inputs.GeoMorph, an extension of DDR, places greater emphasis on multi-dimensional feature registration by introducing a feature extraction network, resulting in better registration performance on both single-dimensional and multi-dimensional feature inputs compared to DDR.SUGAR improves the computational efficiency of the network by applying a graph attention on the sphere. Despite the significant progress in cortical surface registration achieved by deep learning methods, challenges and limitations still exist. Deep learning-based registration methods still fall slightly short of the current state-of-the-art classical method, SD, in terms of registration accuracy. Furthermore, these spherical projection-based methods all employ convolution methods in the spatial domain. Due to the characteristics of spherical data, these convolution methods cannot change the size of the spherical feature maps, and the scaling of spherical feature maps must involve specific spherical upsampling and downsampling methods, which may lead to partial feature loss, thus affecting registration outcomes.

\section{Method}
\label{sec:headings}

\subsection{Overall Framework}
\begin{figure}[htbp]
  \centering
  \includegraphics[width=\textwidth]{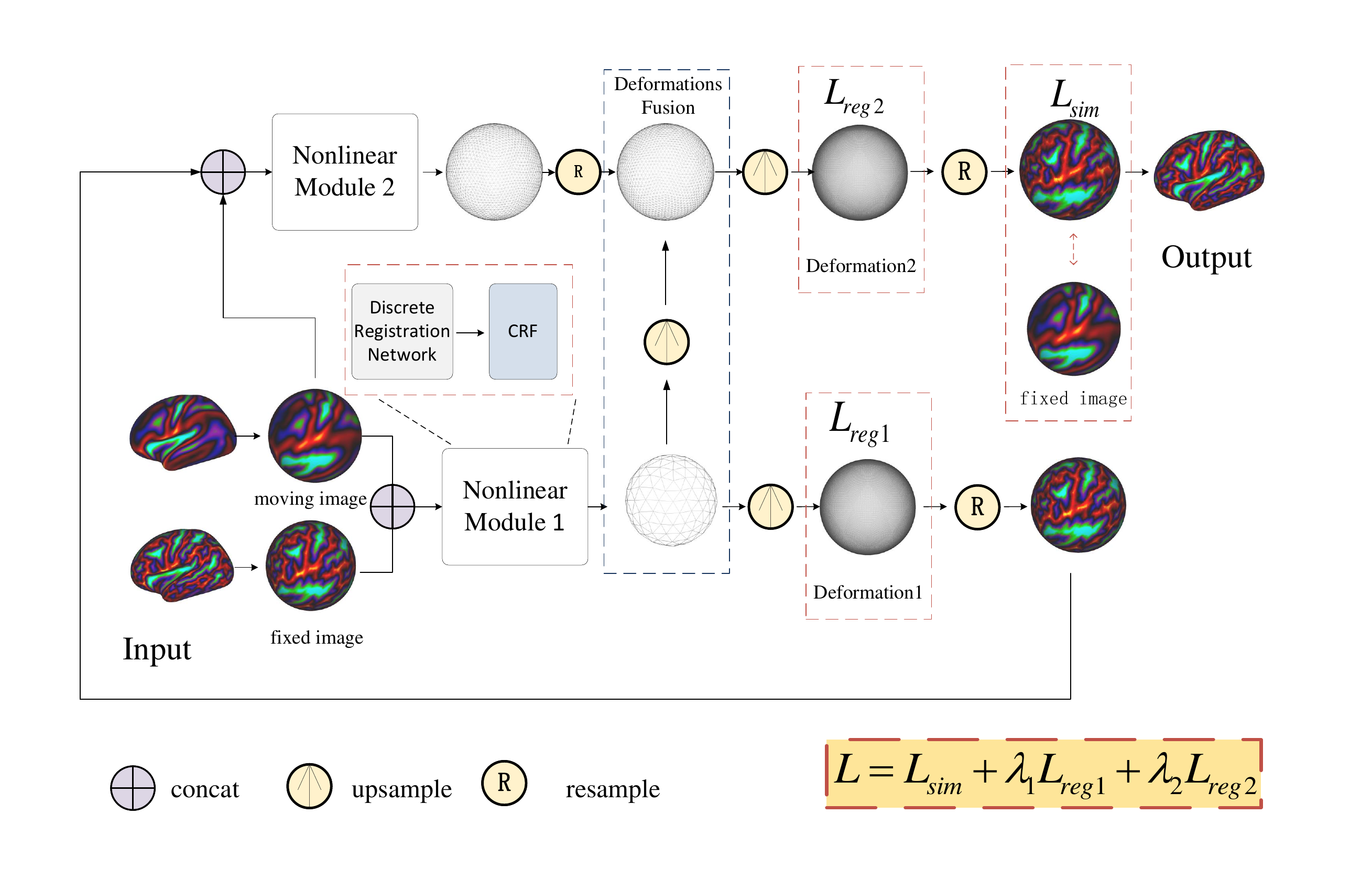}
  \caption{overallframe}
  \label{fig:overall}
\end{figure}

GESH-Net is designed for nonlinear registration, with the overall structure shown in Figure \ref{fig:overall}. Specifically, GESH-Net consists of multiple cascaded nonlinear registration modules at different scales. Each nonlinear registration module at a given scale receives the lower-scale spherical deformation from the previous module, upsampling it to the current scale to perform further registration. The deformations generated at each scale are then fused to produce the final nonlinear registration deformation.
Each nonlinear module is composed of a discrete registration network and a conditional random field. Before performing the nonlinear registration, the moving image and the fixed image are projected onto a icophere and undergo an initial linear alignment. The linearly aligned moving and fixed images are then input into GESH-Net for processing.

\subsection{Discrete registration network}
\begin{figure}[htbp]
  \centering
  \includegraphics[width=0.9\textwidth]{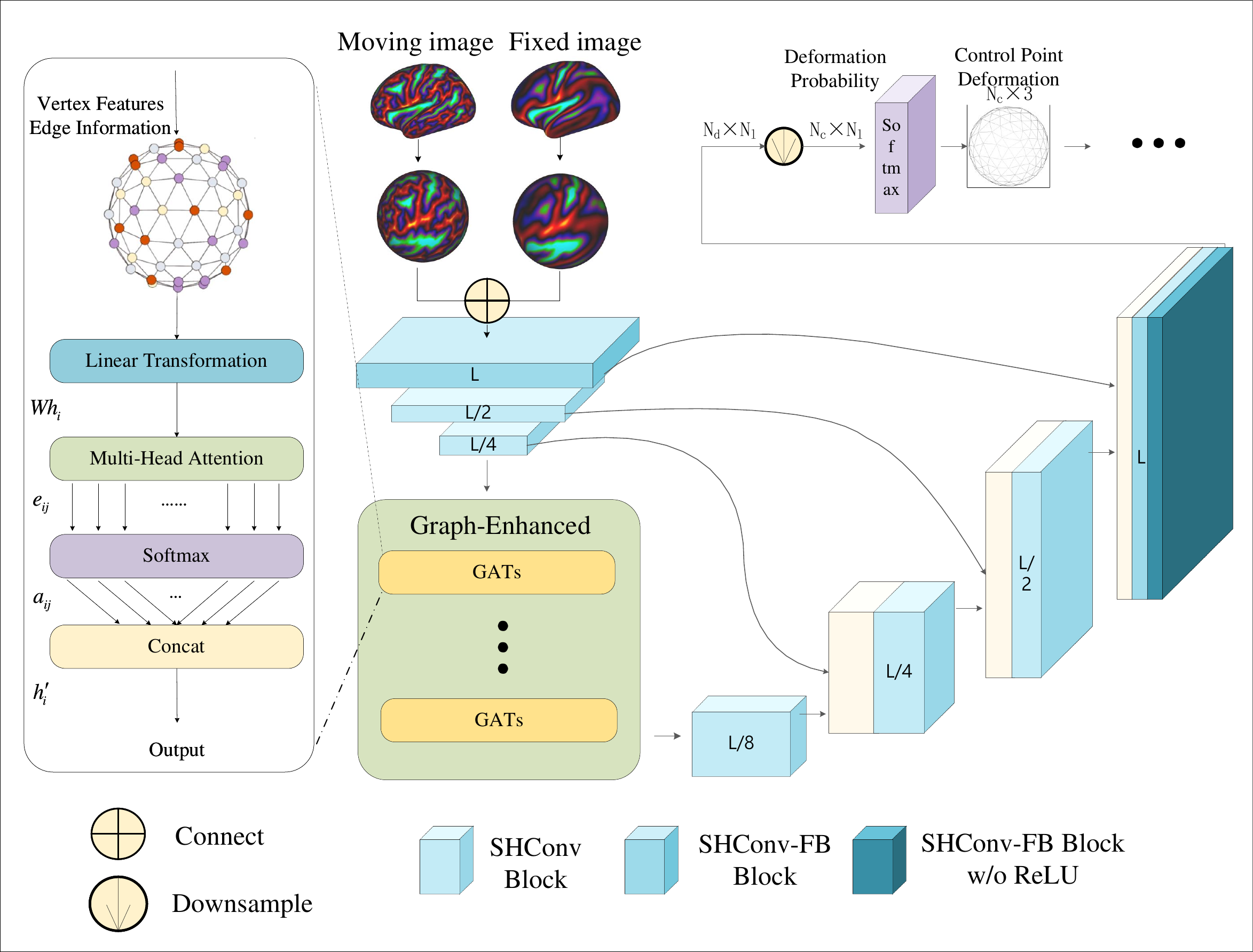}
  \caption{discrete registration network}
  \label{fig:network}
\end{figure}

The discrete registration network includes a U-Net and a Softmax layer, where the U-Net consists of spherical harmonic convolution(SHConv) blocks and a Graph-Enhanced module. This network is used to estimate the probability of control points deforming to nearby label points in the discrete registration method, and its structure is shown in Figure \ref{fig:network}. 
The scaling in the U-Net is based on spectral pooling\cite{Esteves2018}, meaning that the contraction and expansion of the convolutional network are controlled by the bandwidth \( L \) of the spherical harmonic transform rather than by the resolution of the feature maps. The network sets the expansion bandwidth to \( L \) at the input, and in each layer of the encoder, \( L \) is halved while the number of channels \( C \) is doubled, equivalent to downsampling the feature map and increasing the number of channels. In the decoder, \( L \) is doubled, and the number of channels \( C \) is halved, which is equivalent to upsampling the feature map and reducing the number of channels.

The network input is the concatenated feature \( {F_{in}}' \in \mathbb{R}^{N \times 2} \), formed by combining the floating image transformed by the previous module \( {I_m}' \in \mathbb{R}^{N \times 1} \) and the fixed image \( {I_f} \in \mathbb{R}^{N \times 1} \), where \( N \) is the number of points in the input image. After the first SHConv block, the input feature \( {F_{in}}' \) produces an output feature \( {F_1} \in \mathbb{R}^{N \times C} \). \( {F_1} \) is then passed to the second SHConv block, yielding the output feature \( {F_2} \in \mathbb{R}^{N \times 2C} \). After passing through three SHConv blocks, the input feature \( {F_{in}}' \) produces the output feature \( {F_3} \in \mathbb{R}^{N \times 4C} \). The output \( {F_3} \) and the icosahedral sphere \( \Phi \in \mathbb{R}^{N \times 3} \) are then input into the graph attention module. This process continues until the final layer of the U-Net, where the output feature is \( {F_{out}}' \in \mathbb{R}^{N \times N_l} \), where \( N_l \) is the number of label points around each control point as set by the network.

The output \( {F_{out}}' \) is downsampled to the same scale as the control points and passed through the Softmax layer, resulting in the deformation probability prediction \( Q \in \mathbb{R}^{N_c \times N_l} \), where \( N_c \) is the number of control points set by the network. Each control point is moved to the label point with the highest deformation probability, yielding the preliminary deformed control sphere \( D \in \mathbb{R}^{N_c \times 3} \).

\subsection{SHConv block}
\begin{figure}[htbp]
  \centering
  \includegraphics[width=0.8\textwidth]{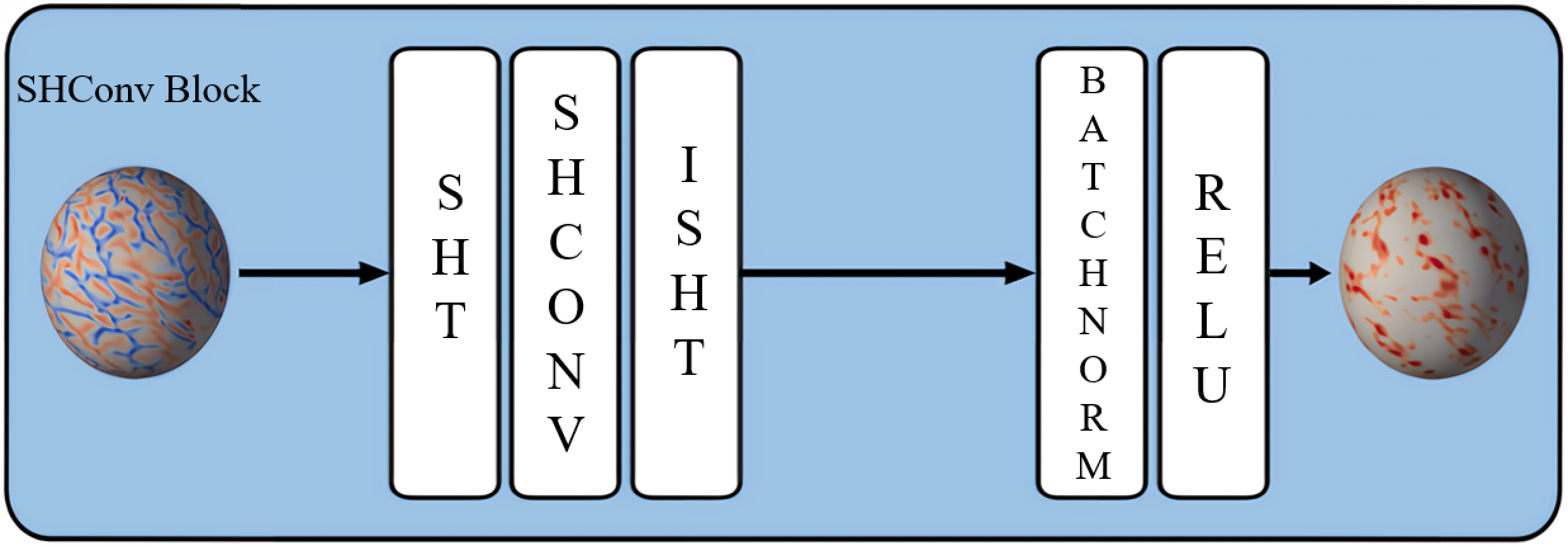}
  \caption{SHConv block}
  \label{fig:SHConv block}
\end{figure}

The SHConv block is a module designed for convolving spherical data based on the spherical harmonic transform and the convolution theorem. It includes operations such as matrix transformation, convolution, activation, and batch normalization. The detailed structure of the SHConv block is shown in Figure \ref{fig:SHConv block}.

In this block, a full-bandwidth SHConv is used to extract high-frequency features by applying a truncation operation on the features. Because in\cite{Ha2022}, high-frequency features can be obtained by multiplying the input features with a learnable parameter. The following sections will detail the algorithmic principles of the SHConv block.

SHConv is a method for convolving features in the spectral domain based on the spherical harmonic transform and the convolution theorem. The convolution filter function on the sphere is defined as \( h(\theta, \phi) \), with its spherical harmonic coefficients given by \( \mathop{h}\limits^\wedge_{lm} \). The convolution theorem states that for two functions defined on the sphere, \( f \) and \( h \), if their spherical harmonic transforms are represented by the spherical harmonic coefficients \( \mathop{f}\limits^\wedge_{lm} \) and \( \mathop{h}\limits^\wedge_{lm} \), respectively, then the spherical harmonic coefficients of their convolution \( (f * h) \), denoted as \( \mathop{(f * h)}\limits^\wedge_{lm} \), can be calculated using Equation \ref{eq:3-1}.
\begin{equation}
    {\mathop {\left( {f * h} \right)}\limits^ \wedge  _{lm}} = \mathop {{f_{lm}}}\limits^ \wedge   * {\mathop h\limits^ \wedge  _{_{lm}}}
    \label{eq:3-1}  
\end{equation}
Since the filter \( h \) is axially symmetric and independent of direction, the convolution operation on the sphere depends only on the degree \( l \) and is independent of the order \( m \). In this case, \( \mathop{h}\limits^\wedge_{lm} = 0 \) for \( m \ne 0 \), meaning that \( \mathop{h}\limits^\wedge_{lm} \) is non-zero only when \( m = 0 \). Therefore, Equation \ref{eq:3-1} can be simplified to the form of Equation \ref{eq:3-2}.
\begin{equation}
    {\mathop {\left( {f * h} \right)}\limits^ \wedge  _{lm}} = C\left( l \right){\mathop f\limits^ \wedge  _{lm}} * {\mathop h\limits^ \wedge  _{_{l0}}}
    \label{eq:3-2}  
\end{equation}
Here, \( C(l) \) is a normalization factor used to ensure the correctness and normalization of the convolution operation in the spectral domain. The SHConv results in the convolved spherical harmonic coefficients. These coefficients are then subjected to the Inverse Spherical Harmonic Transform (ISHT). ISHT is the inverse process of the spherical harmonic transform, and this operation converts the convolution results in the spectral domain, \( \mathop{(f * h)}\limits^\wedge_{lm} \), back into spatial domain features. The transformation formula is given in Equation \ref{eq:3-6}.
\begin{equation}
\left( {f * h} \right)\left( {\theta ,\phi } \right) = \sum\limits_{l = 0}^\infty  {\sum\limits_{m =  - l}^l {C\left( l \right){{\mathop f\limits^ \wedge  }_{lm}} * {{\mathop h\limits^ \wedge  }_{l0}}Y_l^m(\theta ,\phi )} }
\label{eq:3-6}  
\end{equation}
However, due to the limited capacity of machines, it is practically impossible to expand the spherical harmonic functions infinitely in the spherical harmonic transform and inverse spherical harmonic transform. Therefore, existing methods use a band-limited spherical harmonic expansion up to a certain bandwidth \( L \), i.e., expanding Equation \ref{eq:3-6} into the form of Equation \ref{eq:3-7}, and discarding the terms from \( L+1 \) to infinity. The convolution formula with limited bandwidth is given by Equation \ref{eq:3-8}.
\begin{equation}
\left( {f * h} \right)\left( {\theta ,\phi } \right) = \sum\limits_{l = 0}^L {\sum\limits_{m =  - l}^l {C\left( l \right){{\mathop f\limits^ \wedge  }_{lm}} * {{\mathop h\limits^ \wedge  }_{l0}}Y_l^m(\theta ,\phi )} }  + \sum\limits_{l = L + 1}^\infty  {\sum\limits_{m =  - l}^l {C\left( l \right){{\mathop f\limits^ \wedge  }_{lm}} * {{\mathop h\limits^ \wedge  }_{l0}}Y_l^m(\theta ,\phi )} }
\label{eq:3-7}  
\end{equation}
\begin{equation}
\left( {f * h} \right)\left( {\theta ,\phi } \right) = \sum\limits_{l = 0}^L {\sum\limits_{m =  - l}^l {C\left( l \right){{\mathop f\limits^ \wedge  }_{lm}} * {{\mathop h\limits^ \wedge  }_{l0}}Y_l^m(\theta ,\phi )} }
\label{eq:3-8}  
\end{equation}
This approach is more computationally feasible, but such harmonic truncation can discard geometric details, potentially hindering the detection of local features. To address this issue, GESH-Net introduces the full-bandwidth convolution (SHConv-FB) component proposed in [45], which replaces the convolution terms from \( L+1 \) to infinity in Equation \ref{eq:3-7} by using a learnable parameter \( \alpha \) that is multiplied with the input features \( f \). This can be expressed as:
\begin{equation}
\left( {f * h} \right)\left( {\theta ,\phi } \right) = \sum\limits_{l = 0}^L {\sum\limits_{m =  - l}^l {C\left( l \right){{\mathop f\limits^ \wedge  }_{lm}} * {{\mathop {(h}\limits^ \wedge  }_{l0}} - \alpha /C(l))Y_l^m(\theta ,\phi )} }  + \alpha f(\theta ,\phi )
\label{eq:3-9}  
\end{equation}
Here, \( \mathop{h}\limits^\wedge_{l0} \) and \( \alpha \) are learnable parameters. The convolved spherical features \( (f * h)(\theta, \phi) \in \mathbb{R}^{N \times 3} \) undergo subsequent operations such as Batch Normalization (BatchNorm) and non-linear activation (ReLU). At this point, the computation for a single SHConv block is complete.

\subsection{Graph-Enhanced module}
\begin{figure}[htbp]
  \centering
  \includegraphics[width=0.4\textwidth]{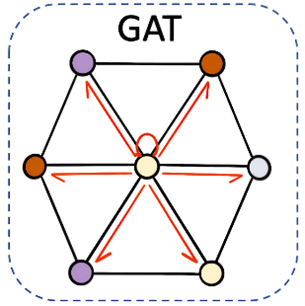}

  \caption{Graph attention in the icosphere}
  \label{fig:GAT}
\end{figure}

\begin{figure}[htbp]
  \centering
  \includegraphics[width=0.8\textwidth]{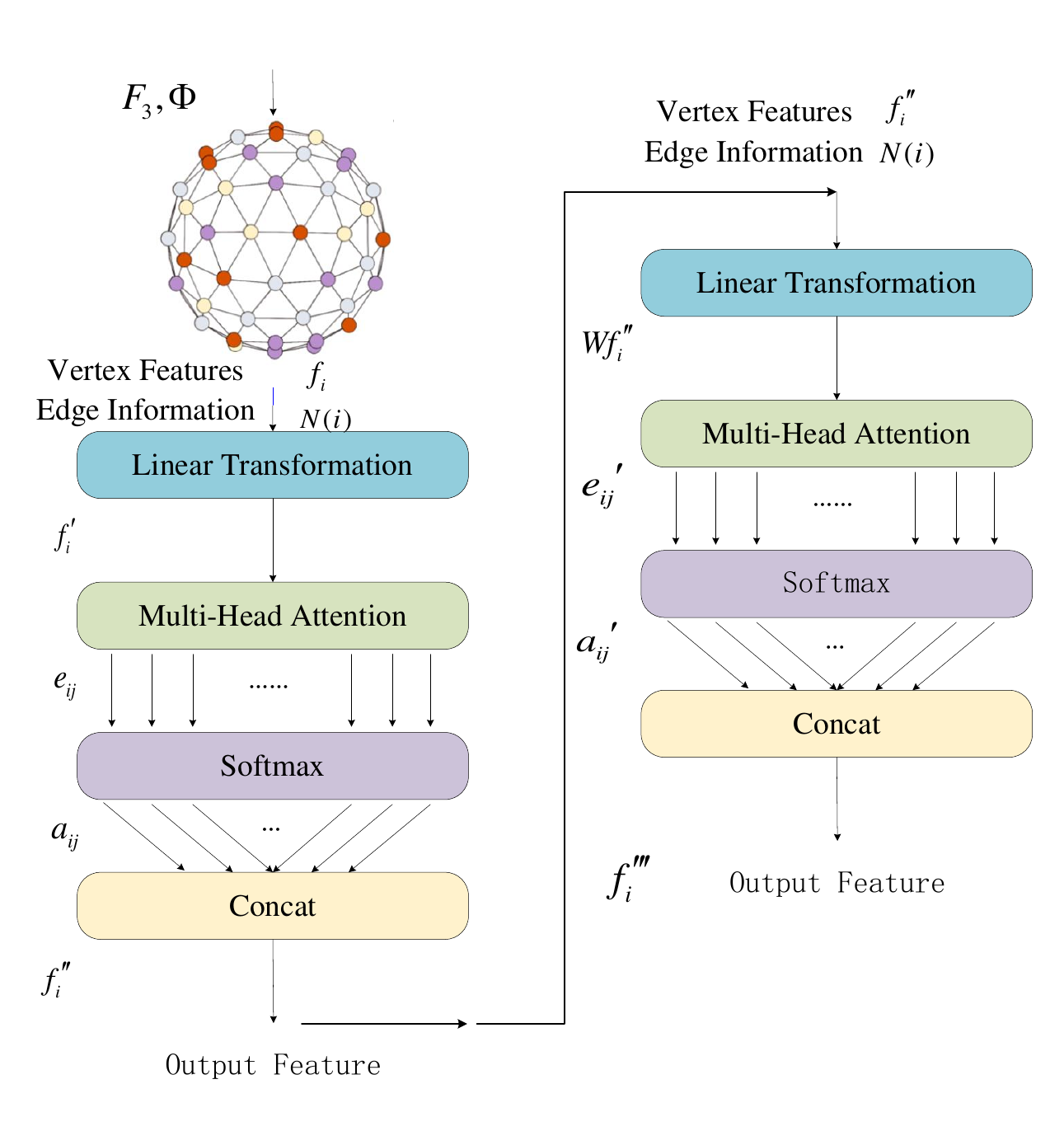}
  \caption{Graph-Enhanced module}
  \label{fig:Graph-Enhanced module}
\end{figure}

The Graph-Enhanced module consists of multiple Graph Attention Networks (GATs), which are neural network architectures designed to work on graph-structured data. In GATs, each node aggregates information from its adjacent nodes with weighted attention, where the weights are dynamically computed through an attention mechanism. This allows the model to emphasize important neighboring nodes and suppress irrelevant ones. In GATs, the input features \( {F_3} \) and \( \Phi \) provide vertex features and edge information, respectively, for the GATs. The weight calculation in the icosphere for GATs is shown in Figure \ref{fig:GAT}. The graph attention module proposed in this chapter consists of two GATs, with the structure illustrated in Figure \ref{fig:Graph-Enhanced module}.
Each GAT consists of four computation steps: linear transformation, attention coefficient calculation, attention coefficient normalization, and weighted feature aggregation. First, the features in \( {F_3} \) undergo a linear transformation. The purpose of the linear transformation is to enable the model to operate in a more appropriate feature dimension, helping to reduce computational complexity and improve representational power. Specifically, a learnable weight matrix \( \mathbf{W} \in \mathbb{R}^{\frac{4C}{H} \times 4C} \) is used, where \( H \) is the number of attention heads set in GATs. For each point \( i \) in \( {F_3} \), its feature value \( f_i \in \mathbb{R}^{4C} \) is transformed into \( f_i \in \mathbb{R}^{\frac{4C}{H}} \), with the calculation formula given by Equation \ref{eq:4-1}.
\begin{equation}
{f_i}^\prime {\rm{  =  W}}{f_i}
\label{eq:4-1}  
\end{equation}
Then, based on the edge information in \( \Phi \), for each adjacent node \( j \) of node \( i \), each attention head calculates an attention coefficient \( e_{ij} \), where \( e_{ij} \) represents the importance of the feature of node \( j \) to node \( i \). This coefficient is computed through a learnable attention mechanism \( a \in \mathbb{R}^{2 \times \frac{4C}{H}} \), which receives the concatenated features of nodes \( i \) and \( j \), and outputs a scalar value. The calculation formula is given by Equation \ref{eq:4-2}.

\begin{equation}
{e_{ij}} = {\rm{LeakyReLU}}({a^T}[{f'_i}||{\mkern 1mu} {f'_j}])
\label{eq:4-2}  
\end{equation}

For ease of comparison and stability, the attention coefficients for each node are normalized using the softmax function. The formula for the normalized attention coefficient \( \alpha_{ij} \) is given by Equation \ref{eq:4-3}.

\begin{equation}
{\alpha _{ij}} = \frac{{\exp ({e_{ij}})}}{{\sum\limits_{k \in N(i)} {\exp } ({e_{ik}})}}
\label{eq:4-3}  
\end{equation}
Here, \( N(i) \) represents the set of neighboring nodes of node \( i \). Finally, each node weights the features of its neighbors according to the calculated attention coefficients, and then aggregates them to update its feature vector \( f \in \mathbb{R}^{4C} \). Since GATs use multi-head attention, \( {f_i}'' \) can be expressed in the form of Equation \ref{eq:4-4}.
\begin{equation}
{f_i}^{\prime \prime } = {\rm{ }}\sum\limits_{h = 1}^H {\left( {\sum\limits_{j \in N(i)} {\alpha _{ij}^{(h)}} {{f'}_{j,h}}} \right)}
\label{eq:4-4}  
\end{equation}
\( \alpha _{ij}^{(h)} \) is the attention coefficient computed by attention head \( h \), and \( {f'_{j,h}} \) is the transformed feature of node \( j \) in the \( h \)-th head. At this point, the computation for a single GAT is complete, and \( {f_i}'' \) forms the output feature \( {F_3}' \in \mathbb{R}^{N \times 4C} \), which serves as the input for subsequent GATs or SHConv blocks.

\subsection{CRF}

\begin{figure}[htbp]
  \centering
  \includegraphics[width=0.8\textwidth]{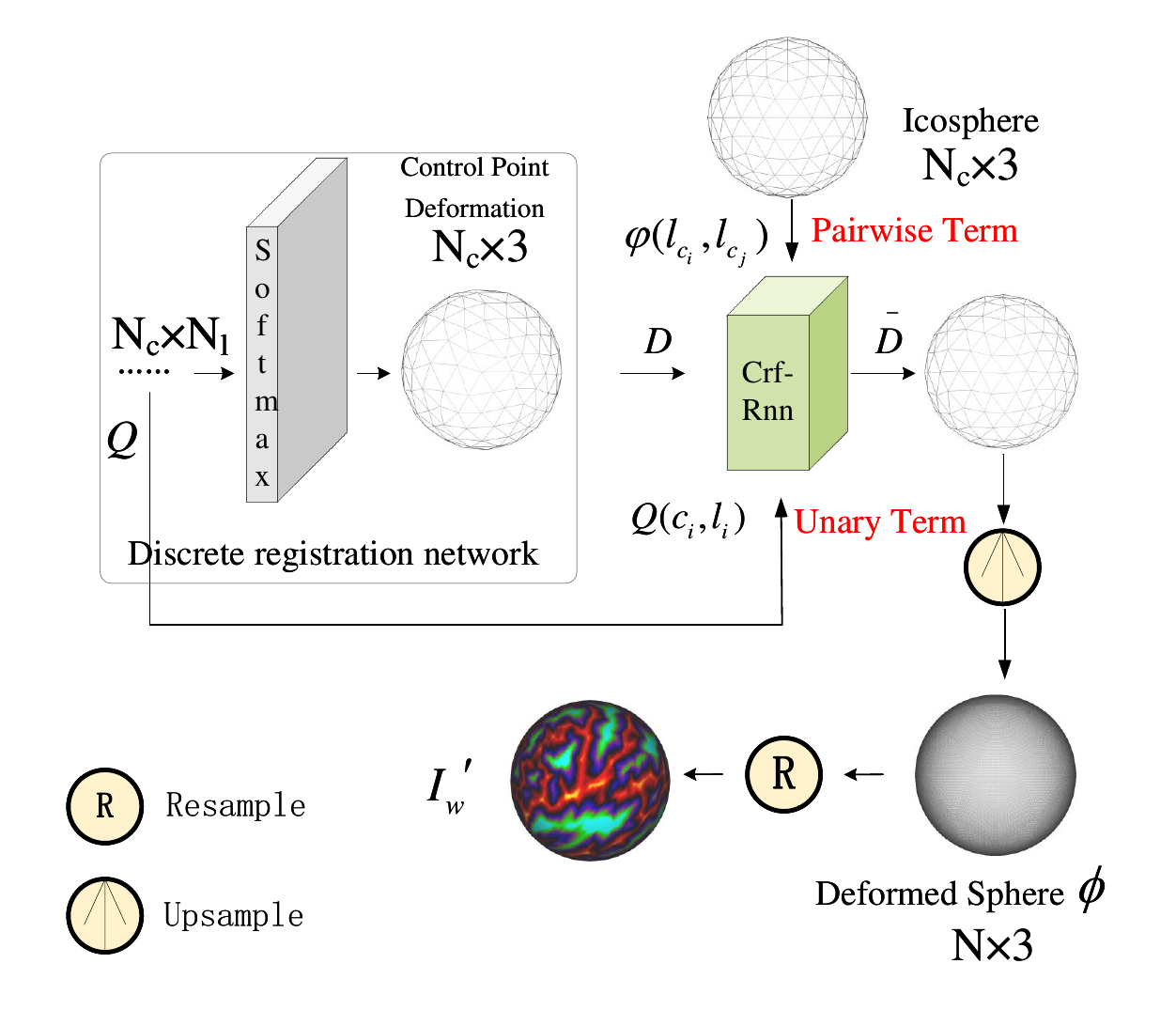}
  \caption{ Conditional Random Field}
  \label{fig: Conditional Random Field}
\end{figure}

The discrete registration network has limited utility because cortical registration is an ill-posed problem with many possible solutions. The deformation \( D \) generated by the discrete registration network model lacks any constraints, allowing each control point to move independently, which may lead to highly distorted deformations. Therefore, this chapter introduces a Conditional Random Field\cite{Sutton2012} [48] (CRF) to impose smoothness constraints on the deformation grid. The CRF is implemented using a Recurrent Neural Network\cite{Grossberg2013} [47] (RNN) and aims to utilize the CRF to constrain the smoothness of the deformation, generating a smoother deformed spherical surface based on the deformation produced by the discrete registration network. The structure is shown in Figure \ref{fig: Conditional Random Field}.
The core concept of the CRF is the energy function, which improves the model's performance by optimizing the energy function. The lower the value of the energy function, the better the compatibility of the configuration with the data, making it more likely to be a good solution. The energy function consists of unary terms and pairwise terms; the unary term represents the energy of a single variable, while the pairwise term represents the energy of the interaction between variables. The energy function used in this chapter is given by Equation \ref{eq:3-14}. 
\begin{equation}
E = \sum\limits_i {Q({c_i},{l_i})}  + \sum\limits_{i \ne j} {\varphi ({l_{{c_i}}},{l_{{c_j}}})}
\label{eq:3-14}  
\end{equation}
Here, the unary term \( Q(c_i, l_i) \) represents the cost of deforming the control point \( c_i \) to the label point \( l_i \), reflecting the cost of moving the control point to a specific label, which is obtained from the probability prediction model shown in Figure \ref{fig:network}. The pairwise term \( \varphi(l_{c_i}, l_{c_j}) \) represents the cost of simultaneously deforming control points \( c_i \) and \( c_j \) to label points \( l_i \) and \( l_j \), and is used to penalize inconsistent deformations of adjacent control points during the deformation process to increase overall smoothness. The pairwise term \( \phi(l_{c_i}, l_{c_j}) \) is optimized by the RNN, with its expression given in Equation \ref{eq:3-15}.

\begin{equation}
\phi ({l_{{c_i}}},{l_{{c_j}}}) = \mu \left( {{l_i},{l_j}} \right){K_G}({l_{{c_i}}},{l_{{c_j}}})
\label{eq:3-15}  
\end{equation}
Here, \( \mu(l_i, l_j) \) is a learnable label compatibility function used to capture the correspondence between different label points, penalizing control points with similar properties that are assigned different labels. This can be obtained from the adjacency relationships of the undeformed standard icosahedron at the same scale as the control points. \( K_G(l_{c_i}, l_{c_j}) \) is a Gaussian kernel function used to compute the spatial similarity between control points. By optimizing the energy function, the deformation optimization network obtains a smoother deformation grid \( \bar{D} \in \mathbb{R}^{N_c \times 3} \). The global optimization capability of the CRF and the iterative learning capability of the RNN effectively increase the smoothness of the discrete registration process, thereby improving the registration accuracy.

\subsection{Optimization}

The entire nonlinear registration network is optimized using an unsupervised loss function, whose form is given in Equation \ref{eq:3-16}. 
\begin{equation}
L = {L_{sim}} + {L_{reg}}
\label{eq:3-16}  
\end{equation}
Here, \( L_{sim} \) measures the similarity between the features on \( F \) and \( M \), and is defined as the sum of the mean squared error and cross-correlation, with the calculation formula provided in Equation \ref{eq:3-17}.
\begin{equation}
{L_{sim}} = {L_{cc}} + {L_{mse}} = \frac{{\sum\limits_{i = 1}^n {({I_{{f_i}}} - {{\bar I}_f})} ({I_{{{w'}_i}}} - {{\bar I}_{w'}})}}{{\sqrt {\sum\limits_{i = 1}^n {{{({I_{{f_i}}} - {{\bar I}_f})}^2}} \sum\limits_{i = 1}^n {{{({I_{{{w'}_i}}} - {{\bar I}_{w'}})}^2}} } }} + \frac{1}{n}\sum\limits_{i = 1}^n {{{({I_{{f_i}}} - {I_{{{w'}_i}}})}^2}}
\label{eq:3-17}  
\end{equation}
\( I_{{f_i}} \) denotes the feature value of the fixed image at vertex \( i \), and \( \bar{I}_f \) represents the average feature value of the fixed image. \( I_{{{w'}_i}} \) indicates the feature value at vertex \( i \) in the final transformed image, and \( \bar{I}_{w'} \) denotes the average feature value of the final transformed image.

The term \( L_{reg} \) represents the regularization loss. For the deformed spherical surface \( \phi \), it is given by \( L_{reg_i} = \lambda_i \left( \left| \nabla \phi_{x_i} \right| + \left| \nabla \phi_{y_i} \right| + \left| \nabla \phi_{z_i} \right| \right) \), where the gradient \( \nabla \) is computed using the hexagonal filter from [29]. We apply regularization penalties to the control grids at two different scales, with different coefficients \( \lambda_i \) for the two \( L_{reg_i} \), as described in Equation \ref{eq:3-18}. 
\begin{equation}
{L_{reg}} = {L_{re{g_1}}} + {L_{re{g_2}}} = \frac{1}{2}\left( {{\lambda _1}\left( {\left| {\nabla {\phi _{{x_1}}}} \right| + \left| {\nabla {\phi _{{y_1}}}} \right| + \left| {\nabla {\phi _{{z_1}}}} \right|} \right) + {\lambda _2}\left( {\left| {\nabla {{\phi '}_{{x_2}}}} \right| + \left| {\nabla {{\phi '}_{{y_2}}}} \right| + \left| {\nabla {{\phi '}_{{z_2}}}} \right|} \right)} \right)
\label{eq:3-18}  
\end{equation}
Here, \( \phi \) is the deformed spherical surface produced by the lower-scale nonlinear module, while \( \phi' \) is the deformed spherical surface produced by the higher-scale nonlinear module.



\section{Experiments}
\label{sec:others}
\subsection{Experimental setup}
\subsubsection{Dataset}
We conducted registration experiments using the S1200 dataset from the Human Connectome Project (HCP)\cite{hcp2012} and the OASIS-1 dataset from the Open Access Series of Imaging Studies (OASIS)\cite{marcus2007open}. 

The HCP S1200 dataset contains high-resolution brain imaging data from 1,200 adult participants. We selected 1,110 sets of cortical surface data from the S1200 dataset for the experiments, dividing them into training, validation, and test sets in an 8-1-1 ratio. The fixed image used the template file provided by HCPpipelines. The acquisition and processing of the HCP S1200 dataset followed strict standardization procedures, ensuring high data quality and consistency.

The OASIS-1 dataset was used for registration experiments to demonstrate the effectiveness of our approach on more complex datasets with smaller sample sizes. OASIS-1 is a publicly available neuroimaging dataset that focuses on brain imaging data of older adults, aimed at studying brain changes related to aging and Alzheimer's disease. It includes data from 436 participants aged
between 18 and 96 years, covering a wide age range. Approximately half of the participants are elderly (age $ \geq 60 $), with some experiencing mild or severe cognitive impairment. We selected 400 sets of cortical surface data from the OASIS-1 dataset for the experiments, also dividing them into training, validation, and test sets in an 8-1-1 ratio. The fixed image used the fsaverage template provided by FreeSurfer.

\subsubsection{Data preprocessing}
The data used in this paper mainly rely on high-resolution 3T MRI scans with T1-weighted imaging. The original MRI images first undergo a series of preprocessing steps, such as motion correction and tissue segmentation, using the FreeSurfer tool. This is followed by surface reconstruction to generate vertices and triangles that describe the geometry of the cerebral cortex, with each hemisphere containing approximately 32,000 vertices. Finally, a series of morphological features, such as sulcal depth and curvature, are computed. These data are then resampled onto a 6th-order icosahedral sphere using barycentric interpolation, resulting in each spherical surface containing 40,962 vertices.For simplicity, registration was driven solely using sulcal depth as a feature, i.e., Cin =2; however, we point out that the network can be straightforwardly adapted to accept multi-channel features.

\subsubsection{Baseline methods}
We validate against the official implementations of SD, MSM Pair, MSM Strain, S3Reg, DDR and GeoMprph. Freesurfer is run as part of the HCP Pipeline, so we obtain its result from there.For the OASIS dataset, we validated against the official implementations of the deep learning methods DDR and GeoMorph.
\subsubsection{Evaluation metrics}
This paper uses the cross-correlation(CC) and distortion metrics between the deformed and fixed grids to evaluate the performance of cortical registration.

The CC represents the similarity between the fixed image \( F \) and the moving image \( M \) after registration. A higher CC indicates a greater similarity between the images before and after registration, with the calculation formula given in Equation \ref{eq:3-19}. Here, \( F_{v_i} \) and \( M_{v_i} \) represent the values at the \( i \)-th point of the fixed and moving images, respectively.
\begin{equation}
CC = \frac{1}{{{N_{\rm{d}}}}}\sum\limits_{i = 1}^{{N_{\rm{d}}}} {\left( {\parallel {F_{{v_i}}} - {M_{{v_i}}}\parallel _2^2} \right)}
\label{eq:3-19}  
\end{equation}
Distortion metrics are used to evaluate the degree of distortion experienced by the triangles on the moving grid during registration. The areal distortion \( J \)\cite{Leopoldo2021} represents the ratio of the triangle's area after deformation to its area before deformation, while the shape distortion \( R \)\cite{Ma2017} indicates the ratio of two edge lengths of a triangle after deformation to their ratio before deformation. The specific calculation method for distortion metrics is as follows: for each triangle vertex \( p, q, r \) on the grid before and after deformation, the local deformation gradient \( F_{pqr} \) is computed. The eigenvalues of \( F_{pqr} \), denoted as \( \lambda_1 \) and \( \lambda_2 \), are obtained using eigenvalue decomposition. These eigenvalues describe the scaling along the two orthogonal eigenvector directions during deformation. The calculation methods for areal distortion \( J \) and shape distortion \( R \) are given by Equations \ref{eq:3-20} and \ref{eq:3-21}.
\begin{equation}
J = {\lambda _1}{\lambda _2}
\label{eq:3-20}  
\end{equation}
\begin{equation}
R = {\lambda _1}/{\lambda _2}
\label{eq:3-21}  
\end{equation}
For convenience in reporting results, we use \( \log_2 \left| J \right| \) to report areal distortion and \( \log_2 \left| R \right| \) to report shape distortion. Smaller values indicate a lower degree of distortion in the deformed grid. It is important to note that as long as the grid is deformed, the distortion value will not be zero. A certain range of distortion is reasonable and necessary for deformation. Therefore, when evaluating distortion, we report not only the mean and maximum values but also the 95th and 98th percentiles of the distortions \( J \) and \( R \) to compare the extent of distortions within the normal range across different methods.

\subsection{Registration result evaluation}

\begin{figure}
    \centering
    \includegraphics[width=0.8\linewidth]{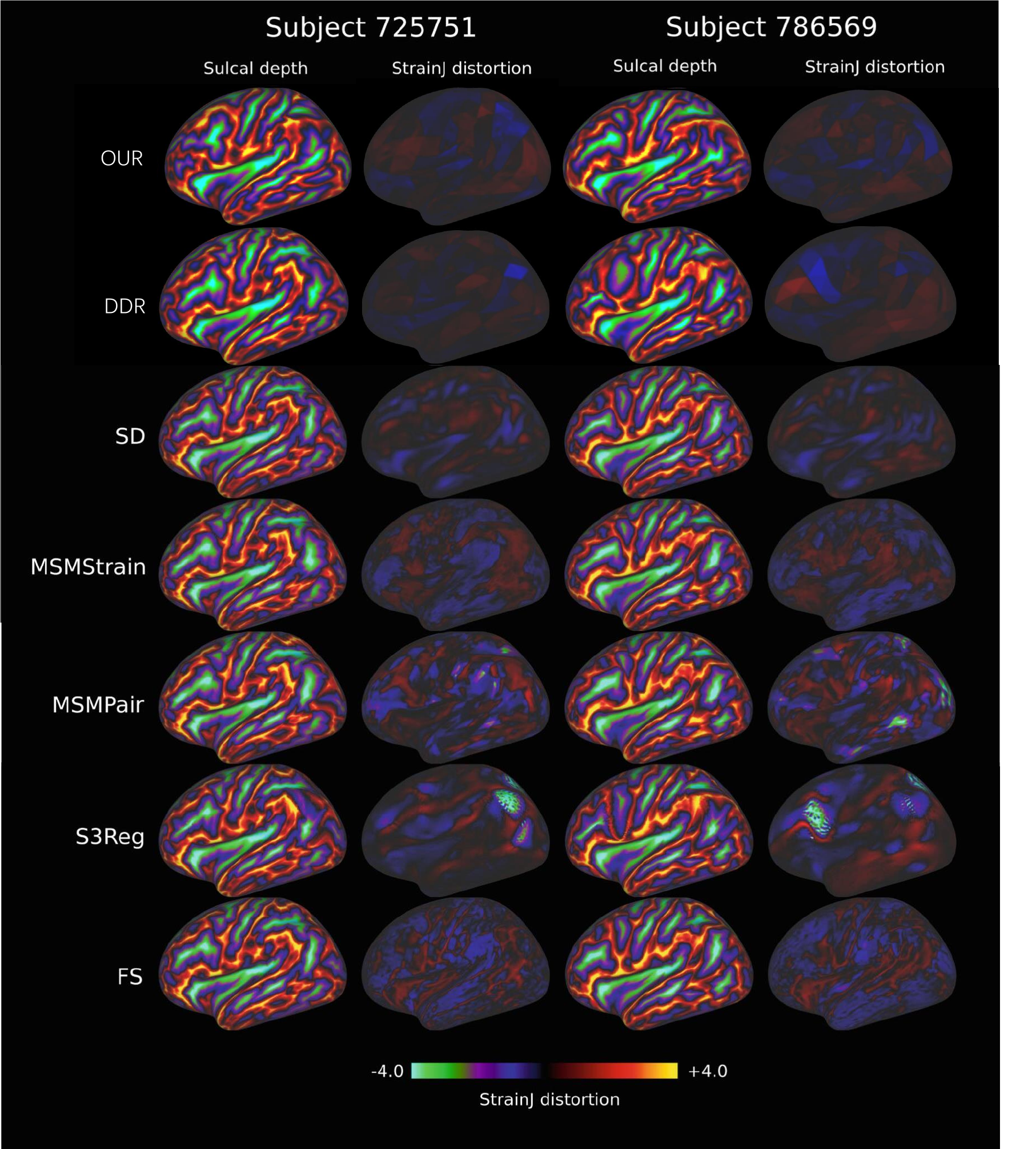}
    \caption{Visualization of Registration Results}
    \label{fig:version_1}
\end{figure}

Figure \ref{fig:version_1} shows the registration results for two test sets from the HCP S1200 test dataset, including the registered sulcal depth features and distortions.
In Figure \ref{fig:version_1}, the distortions for each method are shown in the range of [-4, 4], with dimmer colors representing smaller distortions. The visualization results indicate that the classical method SD, as well as the deep learning methods DDR and our method, produce smaller distortions. The classical methods FS and MSMStrain exhibit generally larger distortion values, while the classical method MSMPair and the deep learning method S3Reg show localized high-brightness areas, representing unreasonable large distortions.

\begin{figure}
    \centering
    \includegraphics[width=0.8\linewidth]{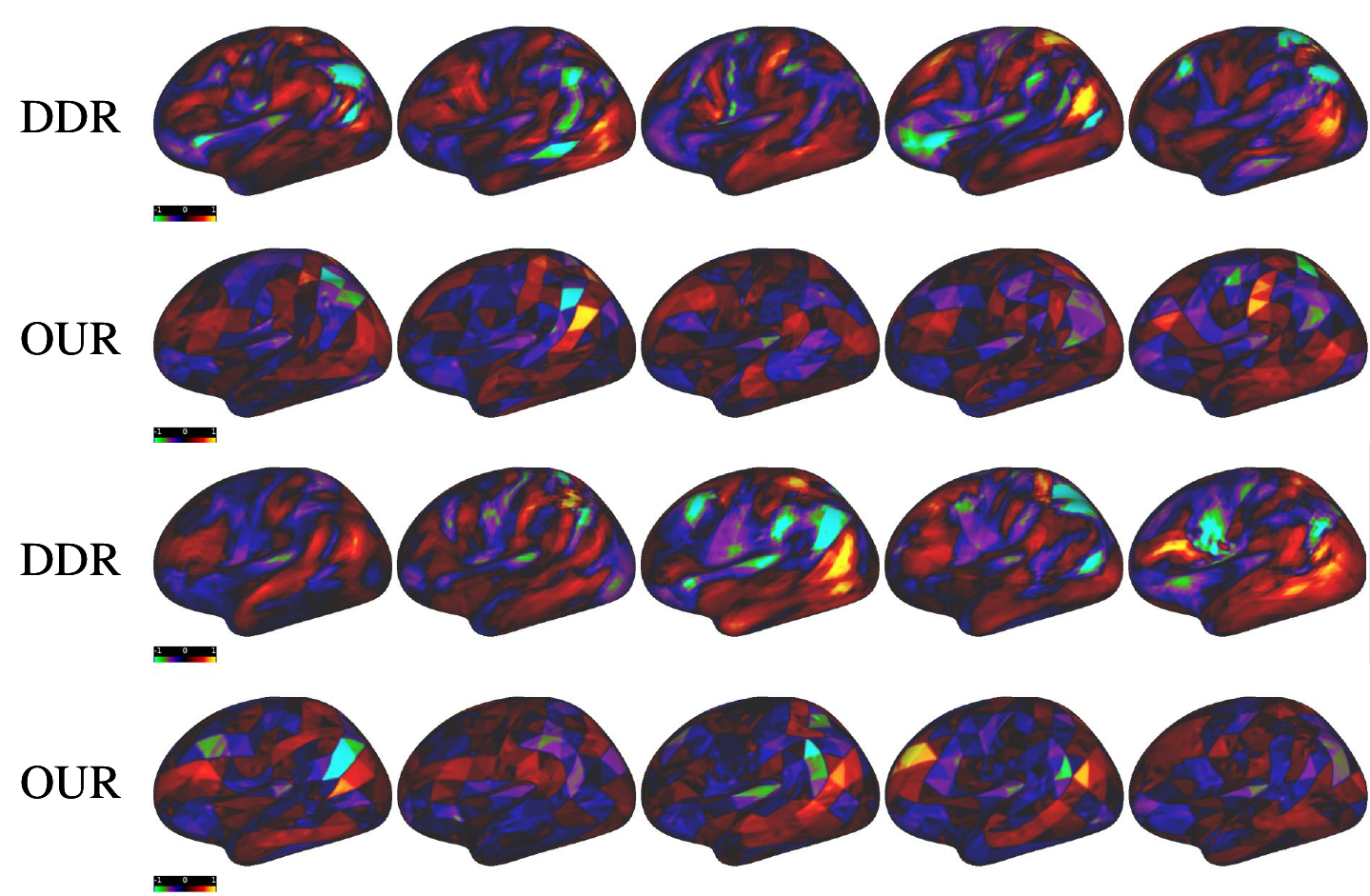}
    \caption{The areal distortions of GESH-Net and DDR}
    \label{fig:version_2}
\end{figure}

\begin{figure}
    \centering
    \includegraphics[width=0.8\linewidth]{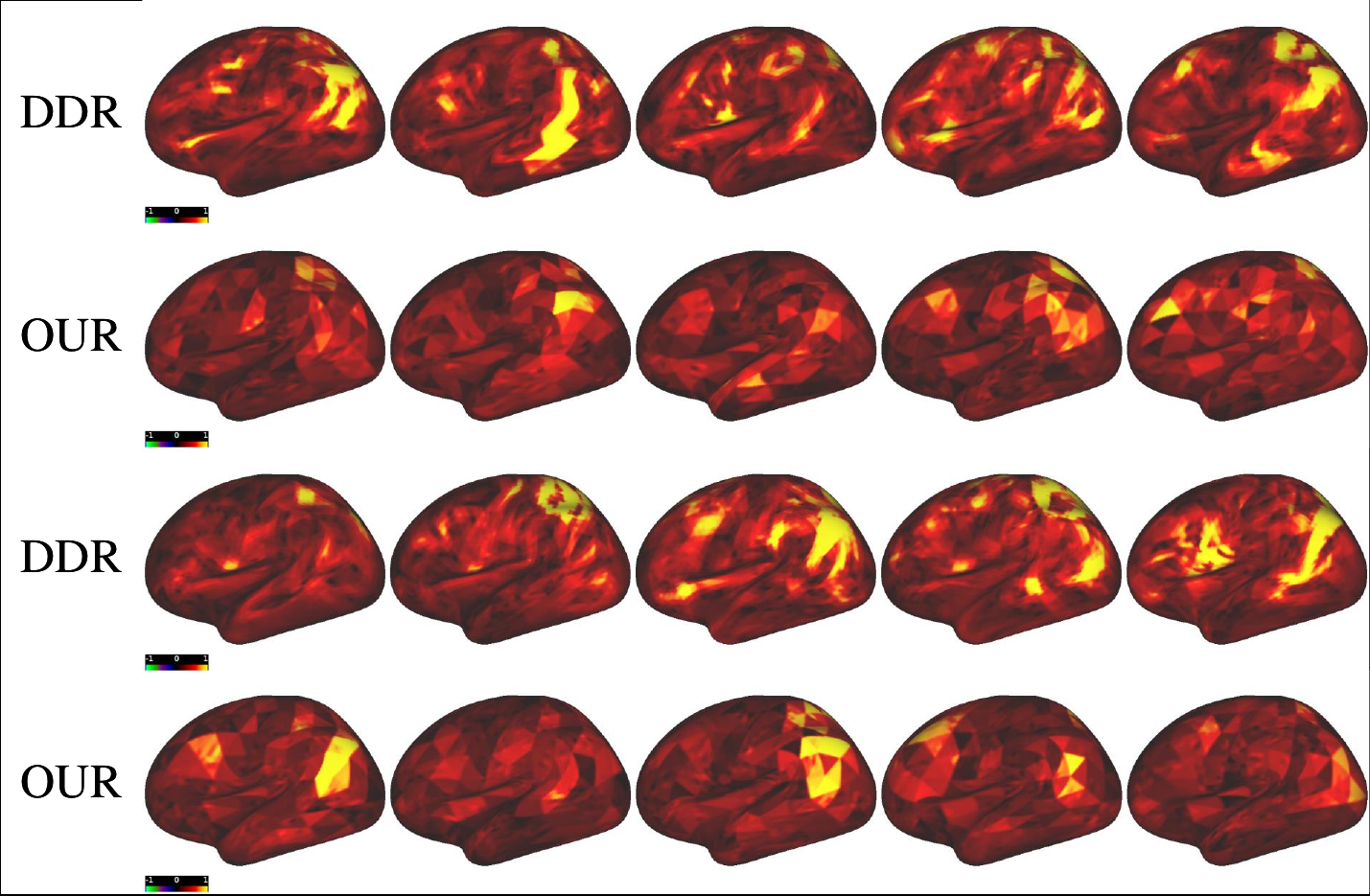}
    \caption{The shape distortions of GESH-Net and DDR}
    \label{fig:version_3}
\end{figure}

Since our method is based on deep learning, and DDR is a key comparison object, we further narrow the color scale to distortions with absolute values less than 1. We select ten consecutive sets of registered images from the test data to compare the registration results of DDR and our method in more detail. The visualizations of areal distortions and shape distortions for DDR and our method are shown in Figures \ref{fig:version_2} and \ref{fig:version_3}.By observing the visualizations of areal distortion and shape distortion, compared to DDR, our method shows significantly fewer high-brightness areas in most selected samples, with the overall color being dimmer. Except for the sixth image, where our method produces more high-brightness areas than DDR, the high-brightness areas in the remaining images are clearly fewer than those in DDR. This indicates that at the same similarity level, the deformations obtained by our method exhibit smaller distortions and smoother deformations.



Table \ref{tab:comparison_hcp} presents the quantitative metrics for the HCP test set, where FreeSurfer is used as a common benchmark registration method. In this chapter, we also report the results using FreeSurfer on the data provided by the HCP.
\begin{table}[htbp]
    \centering
    \caption{Quantitative Comparison of Different Methods}
    \label{tab:comparison_hcp}
    \begin{adjustbox}{max width=\textwidth}
    \begin{tabular}{lccccccccccccc}
        \toprule
        Methods & \multicolumn{2}{c}{CC Similarity} & \multicolumn{5}{c}{Areal Distortion $J$} & \multicolumn{5}{c}{Shape Distortion $R$} & Time \\
        \cmidrule(lr){2-3} \cmidrule(lr){4-8} \cmidrule(lr){9-13}
        & Mean & Std & Mean & Std & Max & 0.95 & 0.98 & Mean & Std & Max & 0.95 & 0.98 & \\
        \midrule
        Freesurfer & 0.815 & 0.0262 & 0.36 & 0.29 & 10.38 & 0.9 & 1.12 & 0.69 & 0.38 & 7.41 & 1.43 & 1.73 & 24min \\
        MSM\_Strain & 0.876 & 0.0197 & 0.26 & 0.33 & 3.04 & 0.51 & 0.63 & 0.56 & 0.36 & 4.91 & 1.04 & 1.21 & 8min \\
        MSM\_Pair & 0.877 & 0.0138 & 0.37 & 0.43 & 9.26 & 1.16 & 1.66 & 0.55 & 0.526 & 8.19 & 1.55 & 2.09 & 42min \\
        SD & 0.875 & (-) & 0.18 & (-) & 2 & 0.5 & 0.65 & 0.24 & (-) & 1.98 & 0.5 & 0.65 & 44s \\
        S3Reg & 0.875 & (-) & 0.266 & (-) & 22.22 & 0.82 & 1.16 & 0.51 & (-) & 21.65 & 1.35 & 2.0 & 11s \\
        DDR & 0.872 & 0.0164 & 0.19 & 0.183 & 7.23 & 0.536 & 0.7 & 0.274 & 0.216 & 7.76 & 0.67 & 0.88 & 3.5s \\
        GeoMorph & 0.873 & 0.0177 & 0.209 & 0.196 & 7.35 & 0.582 & 0.76 & 0.298 & 0.226 & 7.04 & 0.72 & 0.935 & 6.7s \\
        GESH-Net & \textbf{0.875} & \textbf{0.0172} & \textbf{0.154} & \textbf{0.15} & \textbf{2.56} & \textbf{0.43} & \textbf{0.58} & \textbf{0.23} & \textbf{0.175} & \textbf{2.39} & \textbf{0.56} & \textbf{0.74} & \textbf{2s} \\
        \bottomrule
    \end{tabular}
    \end{adjustbox}
\end{table}
From the data in the table, it is evident that our method achieves the smallest distortions at the same similarity level. The mean, 95th percentile, 98th percentile, maximum, and standard deviation of the distortions are significantly better compared to other deep learning methods, and in some metrics, our method even outperforms the current best classical method, SD. Specifically, for the areal distortion metric, our method shows a 14.4\% lower mean than SD, 14\% lower at the 95th percentile, and 10.5\% lower at the 98th percentile. Compared to DDR, the mean is 30\% lower, 27.1\% lower at the 95th percentile, and 25.6\% lower at the 98th percentile. For the shape distortion metric, our method has a 4.1\% lower mean than SD, 12\% higher at the 95th percentile, and 13.8\% higher at the 98th percentile. Compared to DDR, our method shows a 25.8\% lower mean, 24.3\% lower at the 95th percentile, and 20.6\% lower at the 98th percentile. 

To further test the advantages of GESH-Net on smaller and more complex datasets, we conducted registration experiments on the OASIS-1 dataset using the deep learning methods DDR, GeoMorph, and GESH-Net. The experimental results are shown in Table 2. Compared to Table 1, the registration performance of all three methods has decreased. However, compared to other methods, the proposed GESH-Net still achieves the best registration performance.

\begin{table}[htbp]
    \centering
    \caption{Quantitative Comparison of Different Methods}
    \label{tab:comparison_methods}
    \begin{adjustbox}{max width=\textwidth}
    \begin{tabular}{lccccccccccccc}
        \toprule
        Methods & \multicolumn{2}{c}{CC Similarity} & \multicolumn{5}{c}{Areal Distortion $J$} & \multicolumn{5}{c}{Shape Distortion $R$} & Time \\
        \cmidrule(lr){2-3} \cmidrule(lr){4-8} \cmidrule(lr){9-13}
        & Mean & Std & Mean & Std & Max & 0.95 & 0.98 & Mean & Std & Max & 0.95 & 0.98 & \\
        \midrule
        DDR & 0.858 & 0.0198 & 0.18 & 0.184 & \textbf{3.49} & 0.51 & 0.7 & 0.264 & 0.205 & \textbf{4} & 0.64 & 0.83 & 3.5s \\
        GeoMorph & 0.867 & 0.0201 & 0.207 & 0.216 & 10.58 & 0.57 & 0.77 & 0.296 & 0.23 & 8.62 & 0.71 & 0.92 & 6.7s \\
        GESH-Net & \textbf{0.871} & \textbf{0.0172} & \textbf{0.16} & \textbf{0.172} & 5.54 & \textbf{0.45} & \textbf{0.51} & \textbf{0.254} & \textbf{0.196} & 5.87 & \textbf{0.6} & \textbf{0.77} & \textbf{2s} \\
        \bottomrule
    \end{tabular}
    \end{adjustbox}
\end{table}
Additionally, our method is significantly faster than classical methods and also outperforms other deep learning methods in terms of runtime. These quantitative results demonstrate that our method offers better stability, faster computational efficiency, and superior registration performance compared to other methods.

\subsection{Ablation Study}
We conducted ablation experiments on the cascaded network structure, SHConv, and GATs used in GESH-Net to examine how these methods contribute to the improvement of registration performance.
\subsubsection{Ablation of the cascaded network structure}
To compare the effect of the multi-scale cascaded network designed in GESH-Net with the independent coarse-to-fine strategy on registration performance, we fixed the convolution method in the network to SHConv and did not use GATs. We then compared the registration results of using the multi-scale cascaded network versus the independent coarse-to-fine strategy. The independent coarse-to-fine registration applies similarity loss and regularization loss at each scale, while the cascaded network uses the loss function proposed in this chapter (Equation \ref{eq:3-17}). The registration results are shown in Table \ref{tab:strategy_effect}:The results show that the cascaded network designed in this chapter achieves better registration performance.

\begin{table}[htbp]
    \centering
    \caption{Effect of Different Strategies from Coarse to Fine on Matching Accuracy}
    \label{tab:strategy_effect}
    \begin{tabular}{lcccccccccc}
        \toprule
        & \multicolumn{2}{c}{CC} & \multicolumn{4}{c}{Areal Distortion $J$} & \multicolumn{4}{c}{Shape Distortion $R$} \\
        \cmidrule(lr){2-3} \cmidrule(lr){4-7} \cmidrule(lr){8-11}
        Methods & Coarse & Fine & Mean & Max & 0.95 & 0.98 & Mean & Max & 0.95 & 0.98 \\
        \midrule
        Independent CtoF & 0.845 & 0.877 & 0.21 & 5.53 & 0.56 & 0.72 & 0.33 & 6.12 & 0.74 & 0.96 \\
        Cascade Network & 0.83 & 0.877 & 0.21 & \textbf{4.88} & 0.56 & 0.72 & \textbf{0.32} & \textbf{4.63} & \textbf{0.71} & \textbf{0.9} \\
        \bottomrule
    \end{tabular}
\end{table}
\subsubsection{Ablation of SHConv}
To investigate the impact of SHConv used in GESH-Net on registration performance, we did not use the cascaded network or GATs, and instead applied the independent coarse-to-fine strategy. We compared the registration performance of SHConv and MoNet as two convolution methods, with the results shown in Table \ref{tab:convolution_effect}. The results indicate that, under the same network structure, SHConv achieves better registration performance than MoNet.

\begin{table}[htbp]
    \centering
    \caption{Effect of Different Convolution Methods on Matching Accuracy}
    \label{tab:convolution_effect}
    \begin{adjustbox}{max width=\textwidth}
    \begin{tabular}{lccccccccccccc}
        \toprule
        & \multicolumn{2}{c}{CC} & \multicolumn{4}{c}{Areal Distortion $J$} & \multicolumn{4}{c}{Shape Distortion $R$} \\
        \cmidrule(lr){2-3} \cmidrule(lr){4-7} \cmidrule(lr){8-11}
        Methods & Coarse & Fine & Mean & Max & 0.95 & 0.98 & Mean & Max & 0.95 & 0.98 \\
        \midrule
        MoNet & 0.841 & 0.878 & 0.22 & 7.92 & 0.59 & 0.78 & \textbf{0.31} & 7.52 & 0.74 & 0.97 \\
        SHConv & 0.845 & 0.877 & \textbf{0.21} & \textbf{5.53} & \textbf{0.57} & \textbf{0.72} & 0.33 & \textbf{6.12} & 0.74 & \textbf{0.96} \\
        \bottomrule
    \end{tabular}
    \end{adjustbox}
\end{table}
\subsubsection{Ablation of GRAPH-ENHANCED module}
\begin{figure}[htbp]
  \centering
  \includegraphics[width=0.9\textwidth]{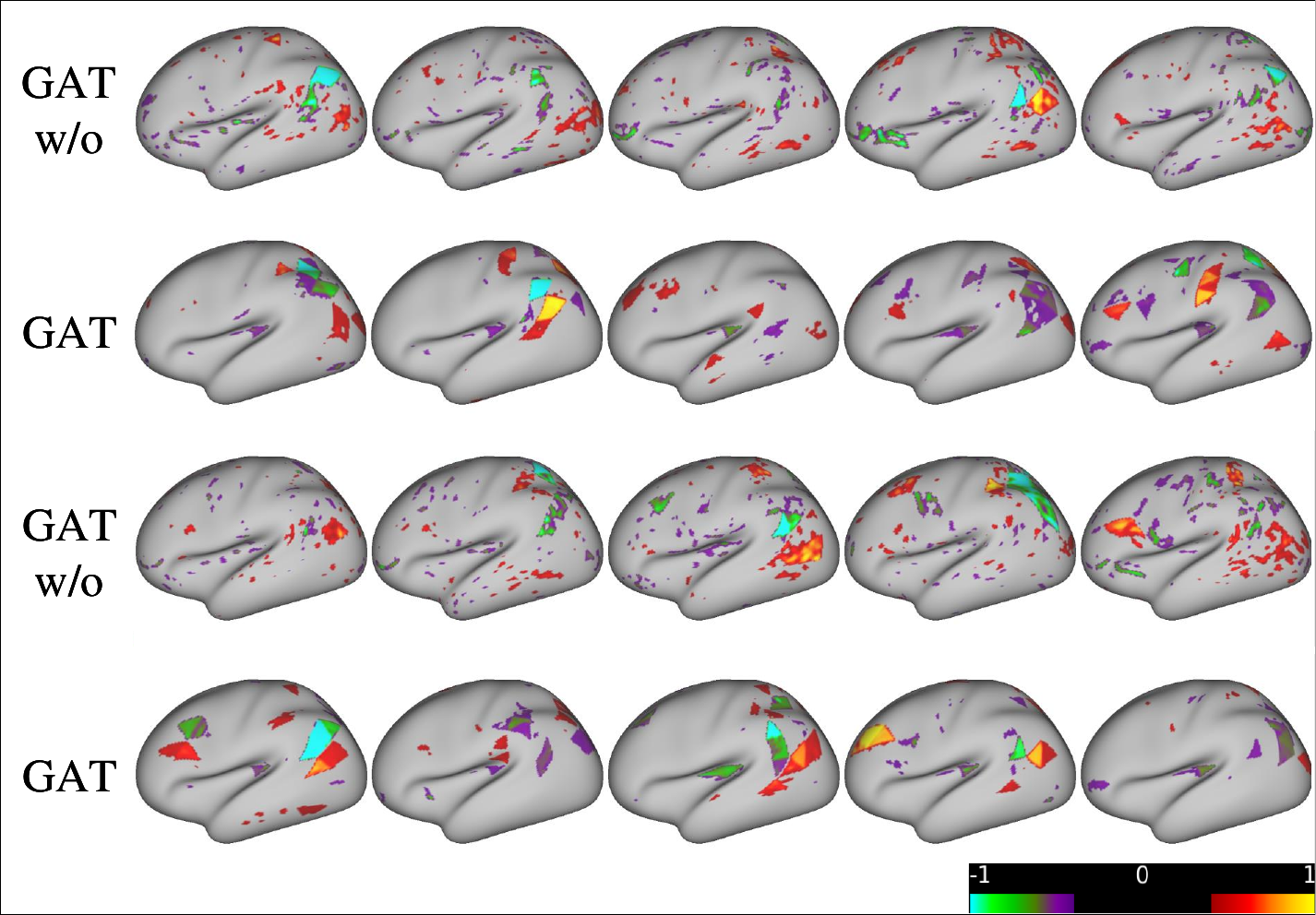}
  \caption{Ablation of GRAPH-ENHANCED J}
  \label{fig:Ablation of GRAPH-ENHANCED J}
\end{figure}

\begin{figure}[htbp]
  \centering
  \includegraphics[width=0.9\textwidth]{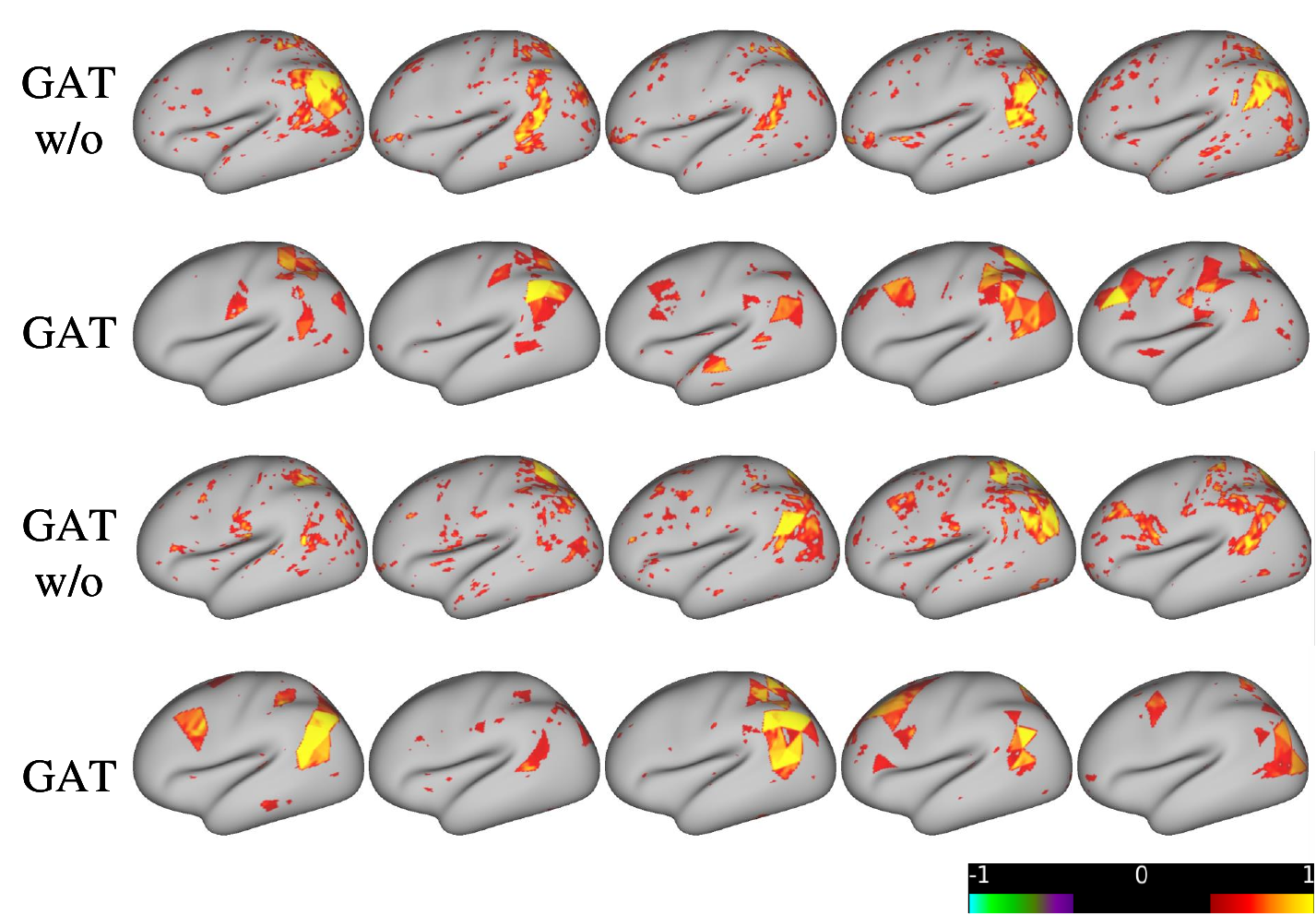}
  \caption{Ablation of GRAPH-ENHANCED R}
  \label{fig:Ablation of GRAPH-ENHANCED R}
\end{figure}

To investigate the impact of the graph enhancement module on the performance of GESH-Net, we performed an ablation study on the graph enhancement module. The quantitative results are shown in Table \ref{tab:gat_ablation}. The ablation results demonstrate that the graph attention module significantly improves the registration performance of the network while only adding a small amount of computation time. To verify the contribution of the graph enhancement module in helping GESH-Net extract global features, we visualized the areas with high distortion values, both areal distortion \( J \) and shape distortion \( R \), after registration on ten sets of test data. We visualized the regions where \( |J| > 0.4 \) and \( |R| > 0.5 \), with the visualization results shown in Figures \ref{fig:Ablation of GRAPH-ENHANCED J} and \ref{fig:Ablation of GRAPH-ENHANCED R}.

\begin{table}[htbp]
    \centering
    \caption{Quantitative Results of Ablation Study on Graph-enhanced Module}
    \label{tab:gat_ablation}
    \begin{adjustbox}{max width=\textwidth}
    \begin{tabular}{lccccccccccccc}
        \toprule
        & \multicolumn{2}{c}{CC Similarity} & \multicolumn{5}{c}{Areal Distortion $J$} & \multicolumn{5}{c}{Shape Distortion $R$} & \\
        \cmidrule(lr){2-3} \cmidrule(lr){4-8} \cmidrule(lr){9-13}
        GATs & Mean & Std & Mean & Std & Max & 0.95 & 0.98 & Mean & Std & Max & 0.95 & 0.98 & Time \\
        \midrule
        w/o & 0.877 & 0.0188 & 0.21 & 0.188 & 4.88 & 0.56 & 0.72 & 0.32 & 0.207 & 4.63 & 0.71 & 0.9 & \textbf{1.9s} \\
        & 0.875 & \textbf{0.0172} & \textbf{0.154} & \textbf{0.15} & \textbf{2.56} & \textbf{0.43} & \textbf{0.58} & \textbf{0.23} & \textbf{0.175} & \textbf{2.39} & \textbf{0.56} & \textbf{0.74} & 2s \\
        \bottomrule
    \end{tabular}
    \end{adjustbox}
\end{table}

\section{Discussion}
Our study introduces a continuous unsupervised learning-based method for cortical surface registration, yielding comparable if not better performance against state-of-the-art conventional and learning-based methods in terms of computational efficiency, registration accuracy, distortion,.These performance benefits are attributed to various aspects of our proposed framework, including the representational capability of the Cascaded structure, the design of the loss, SHConv, and the Graph-enhanced module.These contributions will be discussed in detail in subsequent sections.
\subsection{Cascaded structure and loss}
The cascaded structure in GESH-Net is designed to improve the existing coarse-to-fine methods. Coarse-to-fine approaches play an important role in cortical surface registration. In independent coarse-to-fine methods, the registration method performs registration independently at each scale, aiming to find the best registration at the current scale. Such a strategy achieves local optima at each scale, but the goal of registration is to obtain the optimal deformation at the final target resolution, and multiple local optima may not lead to a global optimum. This phenomenon may be more pronounced in discrete registration. Discrete registration deformations largely depend on the division of label points, as each control point deforms within its corresponding label point. If the optimal point for deformation is not within the range of the label point, the control point cannot achieve the optimal deformation, which is one of the reasons why discrete registration struggles to handle large deformations. The most direct way to solve this problem is to expand the range of label points, but this leads to more conflicts between control points and increases computational complexity.

GESH-Net introduces a cascaded structure to mitigate this issue by training multiple scales of registration modules simultaneously and proposing a loss function tailored for this structure. In our loss function (Equation \ref{eq:3-18}), we apply regularization constraints to the output of all scales, but compute the similarity loss only for the highest scale's registration result. We believe that the role of the lower-scale registration modules is to provide better label point assignments for the higher-scale modules, rather than to achieve the optimal deformation at the lower scales. The results in Table \ref{tab:strategy_effect} also support our hypothesis: the cross-correlation at the coarse stage in GESH-Net is lower than that of the independent coarse-to-fine method, but it achieves better results at the fine stage.
\subsection{Cnvolution method}
SHConv is a convolution method based on the spherical harmonic transform, specifically designed for deep learning on spherical surfaces. The convolution method used in cortical surface registration is not fixed, and due to the structural characteristics of the sphere, convolution methods on the sphere cannot perform scaling of the spherical surface. Typically, scaling of spherical feature maps relies on the methods of icosphere subdivision and refinement, which makes the scaling of spherical feature maps less flexible and difficult to correspond with the scaling of channel numbers. SHConv achieves the effect of feature map scaling by controlling the expansion bandwidth of the spherical harmonic transform, making the design of deep learning frameworks on the sphere more flexible. Additionally, existing work [10] has also validated the effectiveness of SHConv in classification tasks, and the discrete method used in GESH-Net can also be viewed as transforming the registration task into a classification task. Table 10 further demonstrates that SHConv outperforms MoNet in this task.

\subsection{Graph-Enhanced module}
SHConv has limited ability to extract global features on the sphere. Generally, increasing the depth of a U-Net network can improve its ability to extract global features, but experiments in\cite{Ha2022} have shown that increasing the depth of U-Net has a minimal effect on SHConv U-Net. SHConv performs convolution in the frequency domain, and reducing the spherical harmonic expansion bandwidth \( L \) means that fewer spherical harmonics are used to represent spherical features or spherical convolution kernels. In contrast, spatial domain convolution methods (such as 1-Ring convolution\cite{Zhao2021-2}) obtain global features by applying convolution on smaller-sized feature maps after pooling. We believe that the fewer spherical harmonics resulting from a lower bandwidth \( L \) are not equivalent to global features in the spatial domain, which leads to SHConv's weaker ability to extract global features. Additionally, since SHConv does not change the actual size of the feature map, it is very difficult to introduce other convolution methods into the lower layers of U-Net.

Therefore, we designed a graph enhancement module that uses a graph attention mechanism to enhance the network's ability to extract global features. By observing the visualizations in Figures \ref{fig:Ablation of GRAPH-ENHANCED J} and \ref{fig:Ablation of GRAPH-ENHANCED R}, we can see that networks without the graph attention module have a more scattered distribution of high-distortion regions, and the few concentrated areas of distortion all exhibit very high distortion values, indicating the presence of topological errors. This suggests that during deformation, the network tends to produce small deformations in multiple small regions to align the images, which we believe is due to SHConv's insufficient ability to extract global features. After adding the graph attention module, we observed the non-highlighted regions in the figures—those with relatively high but normal distortion values—and found that the distortions include both small scattered regions and large concentrated areas. This demonstrates that the graph attention module improves the network's ability to extract global features while retaining the original network's ability to capture local features.

\section{Conclusion}
In this study, we propose GESH-Net, a solution for non-rigid cortical surface registration. This method demonstrates outstanding performance in terms of computational efficiency, registration accuracy, and deformation control, making it highly effective for surface registration. It holds great potential for advancing large-scale neuroimaging studies. Additionally, the cascaded structure and corresponding loss functions, as well as the incorporation of SHConv and the Graph-Enhanced module, offer a versatile approach that can be extended to various tasks involving spherical mesh representations.
\section*{Acknowledgments}
This was was supported in part by......

\bibliographystyle{unsrt}  
\bibliography{references}

\end{document}